\documentclass[lettersize,journal]{IEEEtran}

\usepackage{amsmath,amsfonts}
\usepackage{algorithm}
\usepackage{algorithmic}
\usepackage{array}
\usepackage[caption=false,font=normalsize,labelfont=sf,textfont=sf]{subfig}
\usepackage{textcomp}
\usepackage{stfloats}
\usepackage{url}
\usepackage{verbatim}
\usepackage{graphicx}
\usepackage{cite}
\usepackage{booktabs}
\usepackage{siunitx}
\usepackage{adjustbox}
\usepackage{tikz}
\usepackage{enumerate}
\usepackage[hidelinks]{hyperref}
\begin{document}

\title{A Bimanual Gesture Interface for ROS-Based Mobile Manipulators Using TinyML and Sensor Fusion}

\author{
    {\large
    Najeeb Ahmed Bhuiyan\textsuperscript{*},
    M. Nasimul Huq\textsuperscript{†},
    Sakib H. Chowdhury\textsuperscript{‡},
    Rahul Mangharam\textsuperscript{§}
    }\\[1ex] 

    \IEEEauthorblockA{\small
        \textsuperscript{*†‡}Department of Mechatronics Engineering, Rajshahi University of Engineering \& Technology,\\
        Kazla, Rajshahi-6204, Bangladesh\\
        \textsuperscript{§}Department of Electrical \& Systems Engineering, School of Engineering \& Applied Science,\\
        University of Pennsylvania, Philadelphia, PA 19104, United States\\
        Email: nabq2000@gmail.com\textsuperscript{*},
               1908009@student.ruet.ac.bd\textsuperscript{†},
               sakib.mte.ruet@gmail.com\textsuperscript{‡},
               rahulm@seas.upenn.edu\textsuperscript{§}
    }
}

\maketitle

\begin{abstract}
Gesture-based control for mobile manipulators faces persistent challenges in reliability, efficiency, and intuitiveness. This paper presents a dual-hand gesture interface that integrates TinyML, spectral analysis, and sensor fusion within a ROS framework to address these limitations. The system uses left-hand tilt and finger flexion, captured using accelerometer and flex sensors, for mobile base navigation, while right-hand IMU signals are processed through spectral analysis and classified by a lightweight neural network. This pipeline enables TinyML-based gesture recognition to control a 7-DOF Kinova Gen3 manipulator. By supporting simultaneous navigation and manipulation, the framework improves efficiency and coordination compared to sequential methods. Key contributions include a bimanual control architecture, real-time low-power gesture recognition, robust multimodal sensor fusion, and a scalable ROS-based implementation. The proposed approach advances Human–Robot Interaction (HRI) for industrial automation, assistive robotics, and hazardous environments, offering a cost-effective, open-source solution with strong potential for real-world deployment and further optimization.
\end{abstract}


\markboth{Preprint}%
{Bhuiyan \MakeLowercase{\textit{et al.}}: A Bimanual Gesture Interface for ROS-Based Mobile Manipulators Using Edge-AI and Sensor Fusion}

\section{Introduction}\label{sec:introduction}
\IEEEPARstart{R}{obot} remote control underpins a wide range of transformative technologies, enabling precise operations across domains like space exploration \cite{spacerobo01, spacerobo02, spacerobo03}, disaster response \cite{disasterrobo01, disasterrobo02}, military operations \cite{militaryrobo01, militaryrobo02}, and assistive robotics \cite{assrobo01, assrobo02}. By harnessing human input to guide robotic systems, this field has unlocked capabilities that extend beyond manual reach, adapting to environments where direct intervention is impractical or hazardous. Gesture-based control emerges as a natural evolution of this concept, leveraging intuitive human movements to command robots with minimal training. This approach proves particularly effective in scenarios requiring immediate, instinctual interaction, such as directing medical neuro-arms with laser scalpels for pinpoint accuracy \cite{neuro_arm} or steering assistive devices to support daily tasks. Its appeal lies in bridging the gap between human intent and robotic action, offering a direct and expressive interface.

Building on this foundation, mobile manipulators—systems that pair a mobile base with a robotic arm—amplify the potential of gesture control, combining locomotion with manipulation for versatile applications. In industrial automation, they streamline assembly lines \cite{momaindustry01, momaindustry02}; in healthcare, they assist with patient care \cite{momahealth01, momahealth02}; and in search-and-rescue missions, they navigate complex terrains \cite{momasearch01}. Their utility extends to extreme conditions: in coal mines, gesture-controlled robots dig under searing heat where workers cannot endure \cite{coal}; in bomb defusing, they shield human operators from lethal risks \cite{bomb}; and in nuclear reactors, they manage radioactive waste without exposing personnel \cite{nuclear}. This synergy of mobility and dexterity makes mobile manipulators a cornerstone of modern robotics. However, gesture-based control, despite its promise, grapples with practical hurdles—unreliable recognition, environmental sensitivity, and resource intensity—that impede its widespread adoption, necessitating innovative solutions.

These limitations arise from fundamental flaws in existing gesture recognition methods, each posing distinct challenges to effective HRI. Vision-based systems, such as Microsoft’s Kinect or Leap Motion, depend on cameras that falter under variable lighting, occlusions, or when users move beyond a narrow range, disrupting gesture detection \cite{visiongesture01, visiongesture02}. Similarly, approaches employing Deep Neural Networks (DNNs) or Convolutional Neural Networks (CNNs) deliver high accuracy but at the cost of substantial power and computational demands, rendering them impractical for lightweight or real-time systems \cite{gestureDNN, gestureCNN}. Electromyography (EMG)-based methods offer an alternative by capturing muscle signals, with studies like one using the Myo Armband achieving 78.94\% accuracy across 10 gestures \cite{stroke_rehab}, yet their reliability wanes as muscle fatigue degrades signal quality over time \cite{emg}. Such challenges are especially pronounced in rehabilitation and mobility assistance, where intuitive, dependable control is paramount. Stroke, a leading cause of disability per the World Stroke Organization \cite{wso_stroke}, often induces hemiparesis, impairing one side of the body, while paralysis afflicts millions globally, severely restricting mobility. Early robot-assisted therapy with patient-driven devices can enhance recovery, and related efforts, such as a Robot Wheelchair (RW) using sensor-based hand gestures via gloves or handles, underscore the demand for accessible interfaces in these contexts \cite{rw_control}.

To address these shortcomings, a more robust system is essential—one that overcomes environmental constraints, reduces resource demands, and maintains consistency across diverse users and conditions. Wearable sensor-based approaches, enhanced by TinyML and edge computing, provide a compelling solution, offering real-time, low-power performance adaptable to dynamic environments \cite{tinymlintro01, tinymlintro02, tinymlintro03}. By integrating sensor fusion—combining data from accelerometers, IMUs, and flex sensors—these systems achieve greater accuracy and robustness, circumventing the pitfalls of vision, neural networks, and EMG \cite{sensefus01, sensefus02}. This paradigm shift not only improves technical feasibility but also aligns with the urgent need for intuitive HRI in rehabilitation, hazardous operations, and beyond, setting the stage for advanced robotic control frameworks.

Inspired by these insights, we propose a dual-hand gesture control system for a ROS-based mobile manipulator, harnessing TinyML and sensor fusion to deliver seamless, efficient operation. The left hand governs the mobile base using an Arduino Nano equipped with an accelerometer and two flex sensors—tilting to command directional movement (forward, backward, left, right) and flexing to fine-tune acceleration or deceleration. Concurrently, the right hand directs the 7-DOF Kinova Gen3 manipulator via an Arduino Nano 33 BLE Sense with TinyML and an LSM9DS1 IMU, mapping distinct gestures: ``Forward-Backward'' to a pickup pose, ``Left-Right'' to placement on the mobile base, ``Flat Rectangle'' to table placement, ``Rectangle'' to elevated placement, ``Circle'' to alternative motion, and ``Up-Down'' to homing. Built on ROS (ros2-jazzy), the system simulates the mobile base in Gazebo Harmonic and controls the manipulator via RViz with MoveIt!. Our primary contributions include:
\begin{enumerate}
    \item \textbf{A dual-hand gesture control architecture} that separates mobile base and manipulator functions, enhancing coordination and natural interaction.
    \item \textbf{A TinyML-based gesture recognition system} for real-time, low-power classification of complex manipulator gestures.
    \item \textbf{A sensor fusion approach} combining accelerometer and flex sensor data for precise mobile base control.
    \item \textbf{A comprehensive ROS-based simulation framework} integrating a custom mobile base and the Kinova Gen3 manipulator.
\end{enumerate}
This framework offers a scalable, cost-effective HRI solution for rehabilitation, mobility assistance, and hazardous operations. The paper proceeds as follows: Section~\ref{sec:related_work} reviews gesture-based control, TinyML, and ROS-based mobile manipulators; Section~\ref{sec:methodology} details the system design, hardware, sensor fusion, and TinyML pipeline; Section~\ref{sec:results} presents experimental results; Section~\ref{sec:discussion} examines findings, limitations, and future directions; and Section~\ref{sec:conclusion} concludes the study.

\section{Related Work}\label{sec:related_work}

Gesture-based robotic control has advanced through several paradigms, including wearable sensor systems, vision-based methods, EMG-driven approaches, and, more recently, TinyML-enabled edge AI. This section surveys key contributions across these domains and positions our dual-hand ROS-integrated system within this evolving landscape.

Systems using IMUs, accelerometers, and flex sensors offer low-latency and direct gesture recognition. Yu et al.~\cite{b1} employed a data glove and CNN-BiLSTM model to control a 5-DOF robotic arm, achieving high accuracy at the cost of high computational load. Simpler systems by Faiz et al.~\cite{b6} and Saleheen et al.~\cite{b7} used IMUs and flex sensors on Arduino platforms but were restricted to single-hand, low-DOF applications. Solly et al.~\cite{b11} used a TinyML model on an Arduino Nano for edge inference with low latency, but with a limited gesture set and no support for bimanual control or ROS integration.

Altayeb~\cite{b2}, Wan Azlan et al.~\cite{b9}, and Mariappan et al.~\cite{b10} employed LSTM, CNN, and stereo vision with accuracies ranging from 96\% to 100\%. However, these methods are sensitive to occlusions and lighting, and their dependence on high-end GPUs limits real-time, mobile use in embedded scenarios.

Muscle-driven gesture systems such as those by Cruz et al.~\cite{b3}, Kim et al.~\cite{b8}, and EunSu et al.~\cite{b12} offer intuitive control but face challenges like signal drift, setup complexity, and fatigue sensitivity. While accurate, they are less scalable and harder to deploy in dynamic or long-term use cases.

Lightweight models for embedded systems have enabled low-power, real-time inference. Nandhini et al.~\cite{b5} used TinyML for 3-DOF control with 95\% accuracy but limited dexterity and gesture variety. Solly et al.~\cite{b11} achieved 2\,ms latency with 5-gesture support, yet lacked bimanual and ROS2 integration. Soori et al.~\cite{b4} explored energy-optimized actuation but without end-to-end gesture control.

\begin{figure}[!h]
    \centering
    \begin{adjustbox}{max width=1.0\columnwidth, center}
        \includegraphics{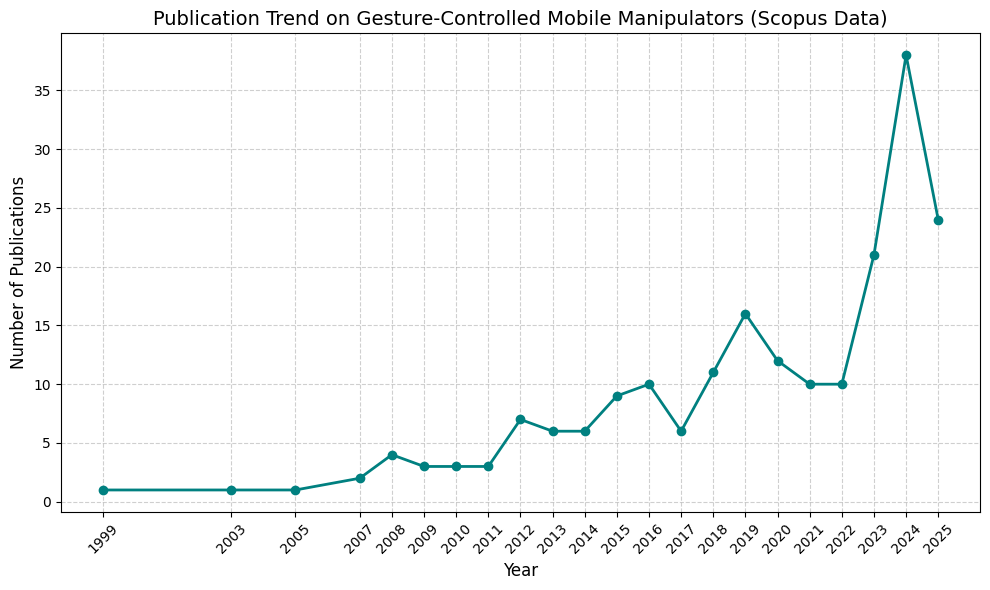}
    \end{adjustbox}
    \caption{Publication trend from Scopus for gesture-controlled mobile manipulators (1999--2025).}
    \label{fig:scopus_trend}
\end{figure}

The increasing research interest is evident from Scopus data, shown in Figure~\ref{fig:scopus_trend}, generated using the query:\\
TITLE-ABS-KEY ( ( "bimanual gesture control" OR "dual-hand gesture" OR "gesture-based control" ) AND ( "mobile manipulator" OR "robotic arm" OR "wheeled robot" ) AND ( "low-cost hardware" OR "embedded system" OR "Arduino" OR "TinyML" OR "ROS2" OR "sensor fusion" ) ) AND PUBYEAR $>$ 1999 AND ( LIMIT-TO ( DOCTYPE , "ar" ) OR LIMIT-TO ( DOCTYPE , "cp" ) ).\\

From 1999 to 2015, research was sparse, reflecting the foundational stage of gesture control in robotics. Growth between 2016 and 2019 coincided with increased access to IMUs and the adoption of ROS. A steep rise from 2020 onward aligns with TinyML breakthroughs, increased demand for contactless systems during the COVID-19 era, and greater investment in assistive and collaborative robotics. With 2024 marking a publication peak, the trend underscores the field’s transition from niche to mainstream.

Our system addresses key limitations across the surveyed works. Unlike single-hand systems~\cite{b1,b6,b7}, our dual-hand architecture separates mobile base and manipulator control, enabling coordinated operation of a 7-DOF Kinova Gen3 arm and a custom mobile platform. Compared to vision-based~\cite{b2,b9,b10} and EMG-based~\cite{b3,b8,b12} methods, our sensor-fusion and TinyML pipeline achieves 99.9\% accuracy with 1\,ms inference latency, robustly operating under varied conditions without external dependencies. Moreover, our ROS2-Gazebo integration provides simulation and deployment scalability absent in many TinyML prototypes~\cite{b5,b11}. Table~\ref{tab:comparative-analysis} presents a comparative summary \footnote{Abbreviations: NR (Not Reported), NS (Not Specified), N/A (Not Applicable), Prost (Prosthetic), Mod (Moderate)}.

Our framework contributes a scalable, real-time, and ROS2-compatible dual-hand gesture interface tailored for mobile manipulators. Its robustness, low-latency, and energy-efficient architecture make it applicable to assistive robotics, human-centric automation, and dynamic industrial environments.

\begin{table}[!h]
    \centering
    \caption{Comparative Analysis of Our Work Against Related Studies}
    \label{tab:comparative-analysis}
    \begin{tabular}{>{\raggedright}p{0.5cm}p{0.5cm}p{1cm}p{1cm}p{1cm}p{1cm}p{1cm}}
        \toprule
        \textbf{Work} & \textbf{DOF} & \textbf{Accuracy (\%)} & \textbf{Latency (ms)} & \textbf{Power Efficiency} & \textbf{ROS Integration} & \textbf{Bimanual Control} \\
        \midrule
        \cite{b1} & 5 & High & NR & Low & No & No \\
        \cite{b2} & 5 & 96 & NR & Low & Yes & No \\
        \cite{b3} & NS & High & NR & Low & No & No \\
        \cite{b4} & NS & N/A & N/A & High & No & No \\
        \cite{b5} & 3 & 95 & Low & High & No & No \\
        \cite{b6} & 3 & NR & NR & Mod & No & No \\
        \cite{b7} & 3 & NR & NR & Mod & No & No \\
        \cite{b8} & Prost & 92 & NR & Mod & No & No \\
        \cite{b9} & 5 & 100 & NR & Low & No & No \\
        \cite{b10} & 6 & 96.54 & High & Low & No & No \\
        \cite{b11} & 5 & High & 2 & High & No & No \\
        \cite{b12} & Prost & 92.5 & Low & High & No & No \\
        \textbf{Ours} & 7 & 99.9 & 1 & High & Yes & Yes \\
        \bottomrule
    \end{tabular}
\end{table}

\section{Methodology}\label{sec:methodology}
This section presents the architectural and algorithmic design of a dual-hand gesture control interface for mobile manipulators. The proposed system consists of: (i) left-hand-based analog control of a mobile base using inertial and flex sensing, and (ii) right-hand-based discrete gesture recognition using a lightweight neural network, along with their respective simulation and visualization frameworks. The design emphasizes modularity, real-time performance, and compatibility with resource-constrained hardware.

\subsection{System Overview}
The bimanual gesture control system comprises two wearable devices for dual-hand operation. The left hand controls the mobile base using an Arduino Nano equipped with an ADXL335 3-axis accelerometer and two flex sensors (SpectraSymbol \(4.5^{\prime\prime}\)). The right hand controls the manipulator via an Arduino Nano 33 BLE Sense, utilizing its embedded LSM9DS1 IMU for gesture recognition. Both devices send serial data or communicate wirelessly via Bluetooth Low Energy (BLE) to a central computer running ROS2 Jazzy, MoveIt2, and Gazebo Harmonic.

The mobile base is simulated in Gazebo, featuring a custom differential-drive platform with two rear wheels and a front caster. The Kinova Gen3 7-DOF manipulator is visualized in RViz. The software stack includes custom ROS~2 nodes for sensor data processing and trajectory execution, Gazebo Harmonic for simulation, MoveIt2 for manipulator planning, and RViz for visualization. This architecture separates control tasks between hands, reducing cognitive load while enabling simultaneous operation of the base and manipulator.

\begin{figure*}[!t]
    \centering
    \includegraphics[width=0.75\textwidth]{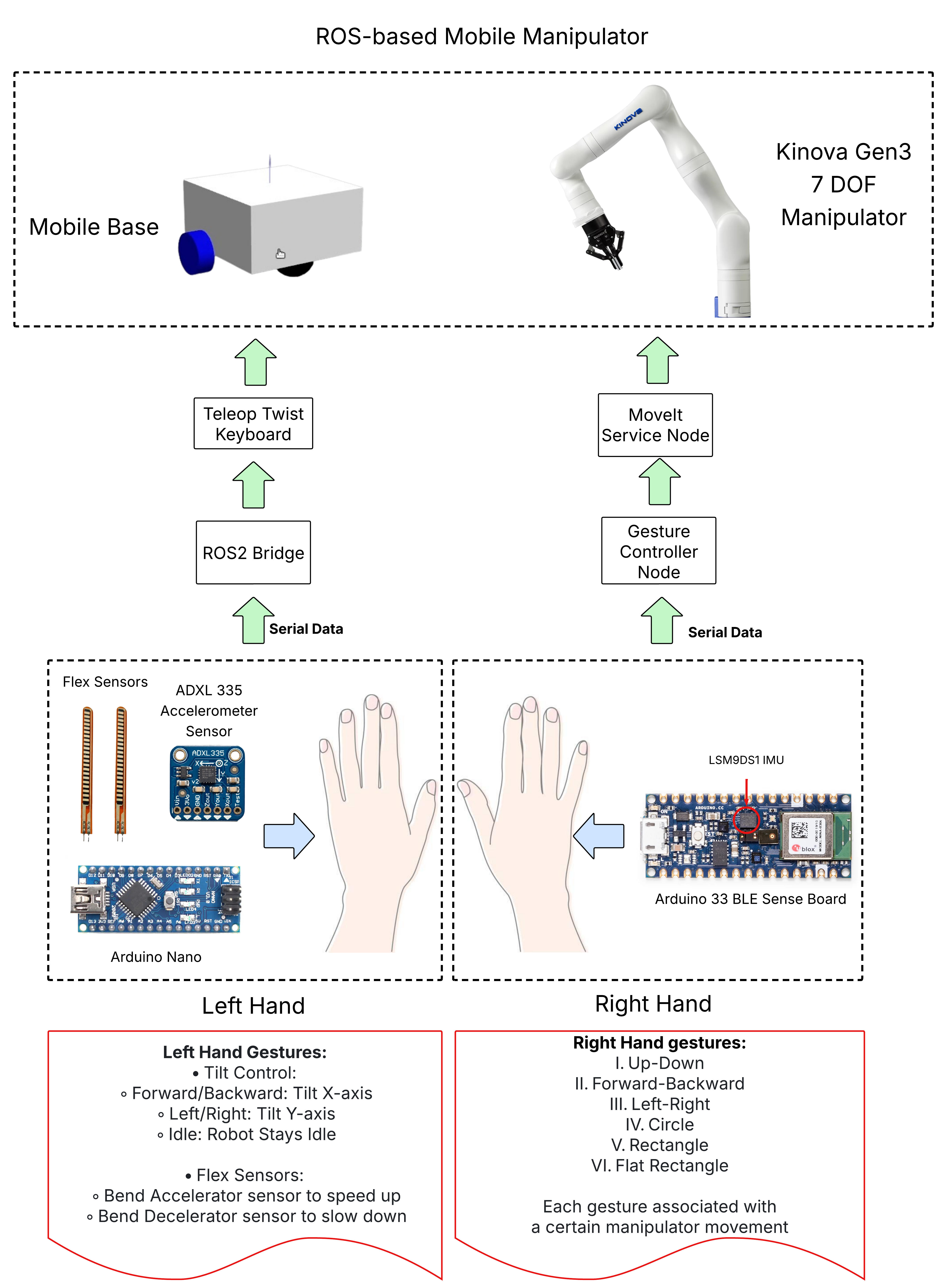}
    \caption{System overview.}
    \label{fig:sys-overview}
\end{figure*}

\subsection{Left-Hand Control for Mobile Base}
The left-hand control system employs sensor fusion to map hand tilt and finger flexion to directional and speed commands for the mobile base, published to the ROS~2 \texttt{/cmd\_vel} topic.

\vspace{0.3cm}

\subsubsection{Accelerometer Tilt Mapping}
The ADXL335 measures acceleration along three axes \((a_x,a_y,a_z)\), from which pitch \((\theta)\) and roll \((\phi)\) are computed \cite{pedley2013tilt}:

\begin{align}
\theta &= \arctan\!\left(\frac{a_x}{\sqrt{a_y^2 + a_z^2}}\right), \label{eq:theta}\\
\phi   &= \arctan\!\left(\frac{a_y}{\sqrt{a_x^2 + a_z^2}}\right). \label{eq:phi}
\end{align}

An analog low-pass filters the accelerometer readings, while dead zones prevent unintended motion: if $|\theta|<5^\circ$ and $|\phi|<5^\circ$, no command is issued. Tilt angles are mapped to discrete commands based on thresholds: forward motion is triggered when $\theta>15^\circ$ ($\text{linear.x}=\text{speed},~\text{angular.z}=0$), backward motion when $\theta<-15^\circ$ ($\text{linear.x}=-\text{speed},~\text{angular.z}=0$), left turn when $\phi>15^\circ$ ($\text{linear.x}=0,~\text{angular.z}=\text{speed}$), and right turn when $\phi<-15^\circ$ ($\text{linear.x}=0,~\text{angular.z}=-\text{speed}$). A calibration step records the neutral hand position.

\vspace{0.3cm}

\subsubsection{Flex Sensor Speed Control}
Two flex sensors on the index and middle fingers are used for acceleration and deceleration respectively. Each detected
bend scales the \emph{maximum} linear and angular speeds by \(10\%\). Define
\[
\gamma_k=
\begin{cases}
1.10, & \text{index bend},\\
0.90, & \text{middle bend}.
\end{cases}
\]
Then both caps are updated by
\begin{equation}
v_{\max} \leftarrow \gamma_k\,v_{\max},\qquad
\omega_{\max} \leftarrow \gamma_k\,\omega_{\max}.
\label{eq:flex_caps}
\end{equation}
The values are limited to preset bounds after the update. The tilt module produces
\((\tilde v(t),\tilde\omega(t))\), which are clipped to the current caps so that
\(|v(t)|\le v_{\max}\) and \(|\omega(t)|\le \omega_{\max}\) \cite{saggio2015resistive}.
Calibration sets sensor thresholds and the initial caps.

Table~\ref{tab:gesture_mapping} shows the gesture to robot mapping for controlling the direction and speed of the mobile base.

\vspace{0.3cm}

\subsubsection{Mobile Base Simulation}
\begin{figure}[!t]
    \centering
    \includegraphics[width=0.9\linewidth]{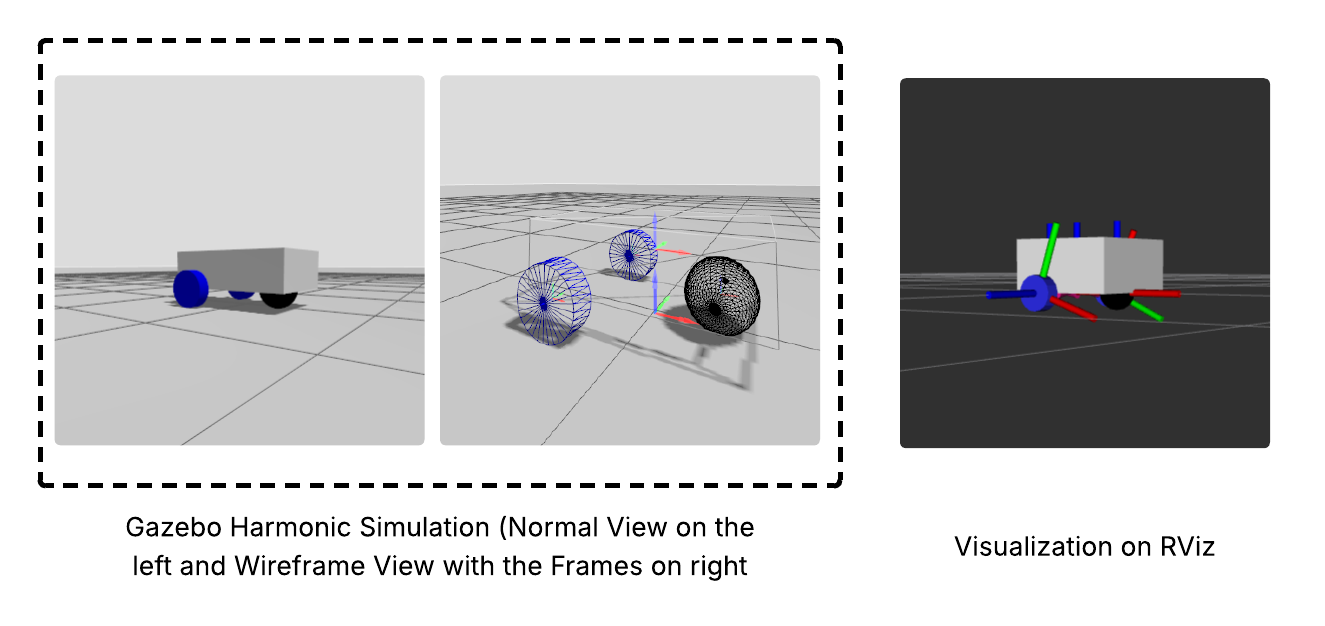}
    \caption{Mobile base in Gazebo Harmonic and RViz.}
    \label{fig:mobile-base-gz}
\end{figure}

The mobile base is controlled in a ROS2–Gazebo environment. Fig.~\ref{fig:mobile-base-gz} shows the base in Gazebo Harmonic and RViz.

A launch file sets up the simulation environment, instantiates the robot model, and establishes communication between the middleware and the simulator. The robot description defines the main body, drive joints, and auxiliary components, with appropriate inertial and dynamic parameters. Control plugins subscribe to velocity commands and provide odometry and transform information, while joint states are made available for visualization. Physical interaction parameters, such as friction, are tuned for stable and realistic motion.

The wearable interface integrates inertial and flex sensing for directional and speed control. Sensor data is preprocessed with thresholding and filtering to reduce noise, then transmitted to the host system. A python script translates the parsed signals into motion commands, which are published to the control topic. Middleware bridges ensure compatibility between message formats used by the control framework and the simulator.

\begin{table}[!t]
    \centering
    \caption{Gesture to Robot Motion Mapping}
    \label{tab:gesture_mapping}
    \begin{tabular}{>{\raggedright}p{1.2cm}p{1cm}p{1.5cm}p{3cm}}
        \toprule
        \textbf{Sensor Input} & \textbf{Keyboard} & \textbf{Robot Motion} & \textbf{Twist Values (\si{m/s}, \si{rad/s})} \\
        \midrule
        X-Axis Tilt Forward & Forward & Move forward & \verb|linear.x = +0.5|, \verb|angular.z = 0| \\
        \addlinespace
        X-Axis Tilt Backward & Backward & Move backward & \verb|linear.x = -0.5|, \verb|angular.z = 0| \\
        \addlinespace
        Y-Axis Tilt Left & Left & Rotate CCW & \verb|linear.x = 0|, \verb|angular.z = +0.5| \\
        \addlinespace
        Y-Axis Tilt Right & Right & Rotate CW & \verb|linear.x = 0|, \verb|angular.z = -0.5| \\
        \addlinespace
        No Tilt (Neutral) & Stop & Halt & \verb|linear.x = 0|, \verb|angular.z = 0| \\
        \addlinespace
        Flex Sensor 1 Bend & Increase max speed & Scale up & \verb|1.10 x linear.x|, \verb|1.10 x angular.z|\\
        \addlinespace
        Flex Sensor 2 Bend & Decrease max speed & Scale down & \verb|0.90 x linear.x|, \verb|0.90 x angular.z| \\
        \bottomrule
    \end{tabular}
\end{table}

Key challenges included message bridging and physics tuning; these were addressed via \texttt{ros\_gz\_bridge}, thresholding/debouncing, and inertial/friction calibration.

\subsection{Right-Hand Control for Manipulator}
The right-hand control uses TinyML for real-time gesture recognition, triggering predefined MoveIt2 trajectories.

\vspace{0.3cm}

\subsubsection{Gesture Dataset Creation}
\begin{figure*}[!t]
    \centering
    \includegraphics[width=\textwidth]{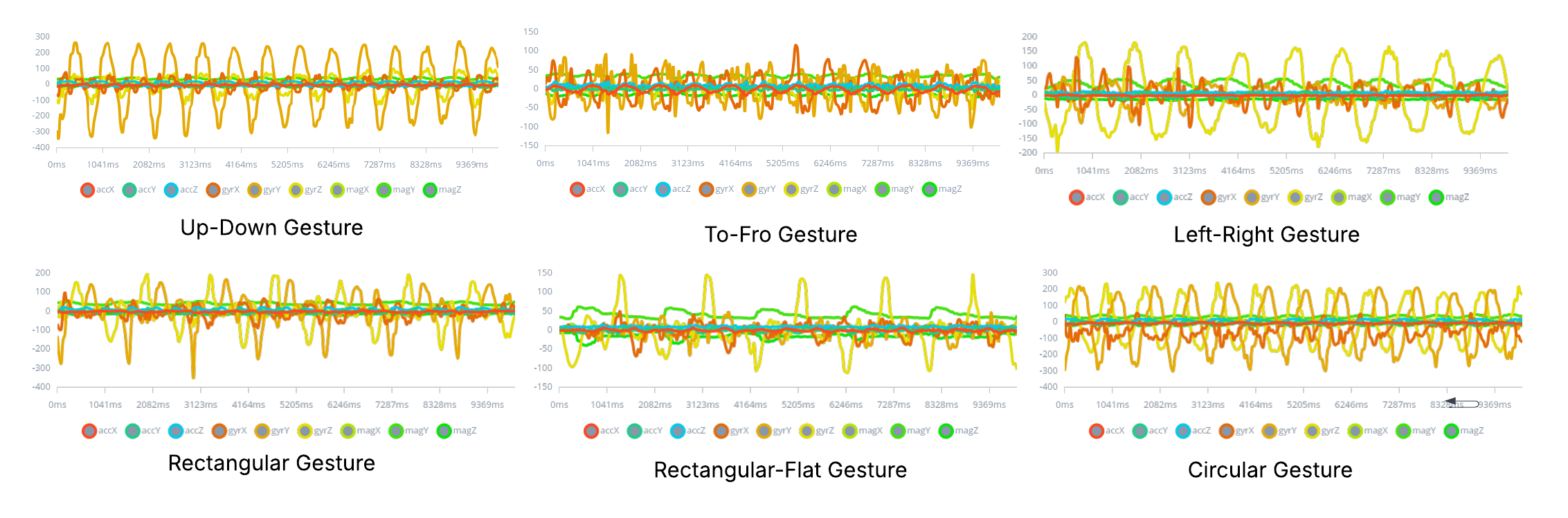}
    \caption{Raw IMU data across gesture classes.}
    \label{fig:raw-data}
\end{figure*}

A custom dataset was recorded using the LSM9DS1 IMU onboard the Arduino Nano~33 BLE Sense. Seven gesture classes (including an \textit{idle} class) were captured, with each gesture performed for approximately \(2~\text{min}~30~\text{s}\) at a sampling frequency of \(f_s=100~\text{Hz}\). For every class, nine IMU axes were collected: accelerometer \((\text{accX}, \text{accY}, \text{accZ})\), gyroscope \((\text{gyrX}, \text{gyrY}, \text{gyrZ})\), and magnetometer \((\text{magX}, \text{magY}, \text{magZ})\). A temporal window of length \(T=2000~\text{ms}\) was defined, yielding:  

\begin{equation}
N = f_s \times T = 100 \times 2 = 200,
\end{equation}

\noindent samples per window. Across all seven classes, the total labeled dataset duration was \(21~\text{min}~11~\text{s}\). Fig.~\ref{fig:raw-data} illustrates representative raw signals.  

\subsubsection{FFT Spectral Analysis}
Each window underwent hybrid feature extraction, combining time-domain statistics with frequency-domain spectral features. Prior to transformation, basic filtering was applied with a scale factor of~1 and an input decimation ratio of~1 to normalize the signals without altering their content. A 16-point FFT was then applied on each axis to provide low-cost spectral content:

\begin{equation}
X[k] = \sum_{n=0}^{N-1} x[n] e^{-j \tfrac{2\pi}{N}kn}, 
\quad k = 0,1,\ldots,N-1.
\end{equation}

The corresponding frequency resolution is:  

\begin{equation}
\Delta f = \frac{f_s}{N_{\mathrm{FFT}}} = \frac{100}{16} = 6.25~\text{Hz},
\end{equation}

\noindent with a maximum frequency component of:  

\begin{equation}
f_{\max} = \left(\frac{N_{\mathrm{FFT}}}{2} - 1\right) \Delta f 
= 43.75~\text{Hz}.
\end{equation}

The power spectral density (PSD) was computed as:  

\begin{equation}
P[k] = \frac{1}{N}\lvert X[k] \rvert^2, 
\quad k = 0,1,\ldots,\tfrac{N_{\mathrm{FFT}}}{2}-1.
\end{equation}

From each axis, eight spectral features and five statistical time-domain features, like mean, root mean square, variance, skewness and kurtosis, were extracted, resulting in a total of:  

\begin{equation}
\text{Features per window} = 9 \times 13 = 117.
\end{equation}

\begin{figure*}[!t]
    \centering
    \includegraphics[width=\textwidth]{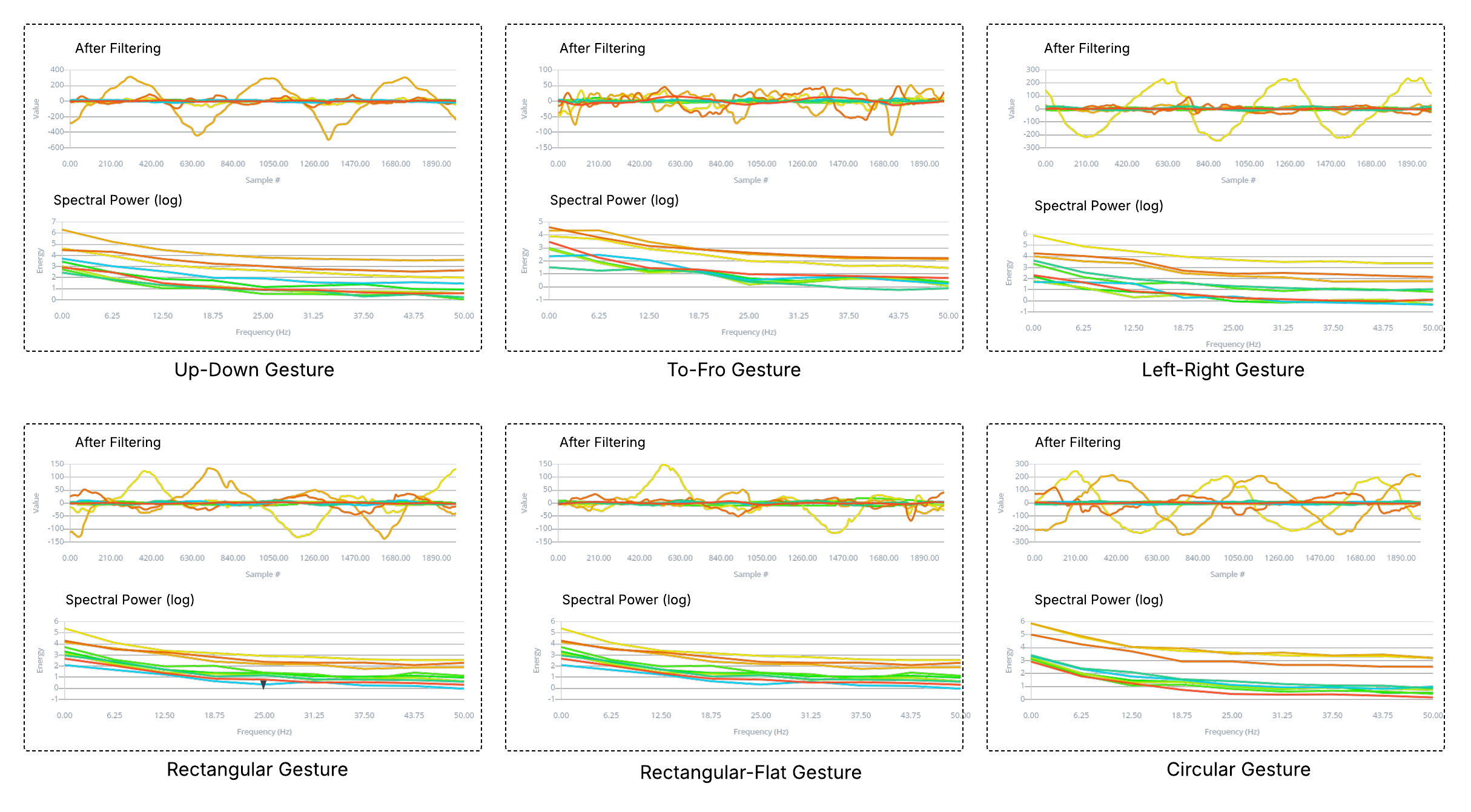}
    \caption{Filtered IMU data and log-scaled spectral power after FFT.}
    \label{fig:fft-pipeline}
\end{figure*}

\begin{figure}[!t]
    \centering
    \includegraphics[width=\linewidth]{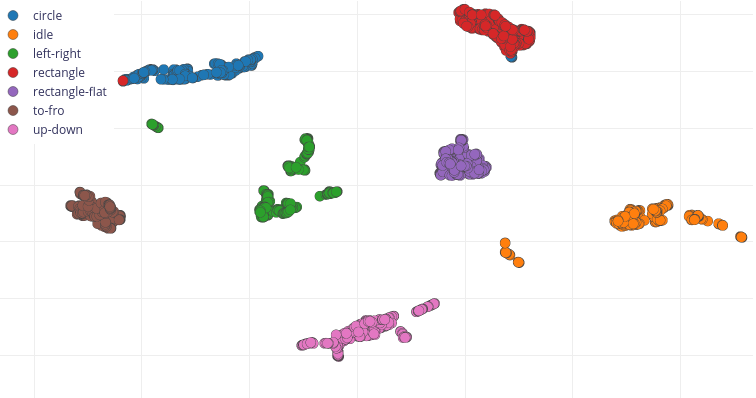}
    \caption{Extracted features from FFT spectral analysis.}
    \label{fig:extracted-features}
\end{figure}

Figs.~\ref{fig:fft-pipeline} and \ref{fig:extracted-features} present the filtered spectra and extracted feature representations, respectively.  

\vspace{0.3cm}

\subsubsection{TinyML Pipeline}
\begin{figure*}[!t]
    \centering
    \begin{adjustbox}{max width=\textwidth, center}
        \includegraphics{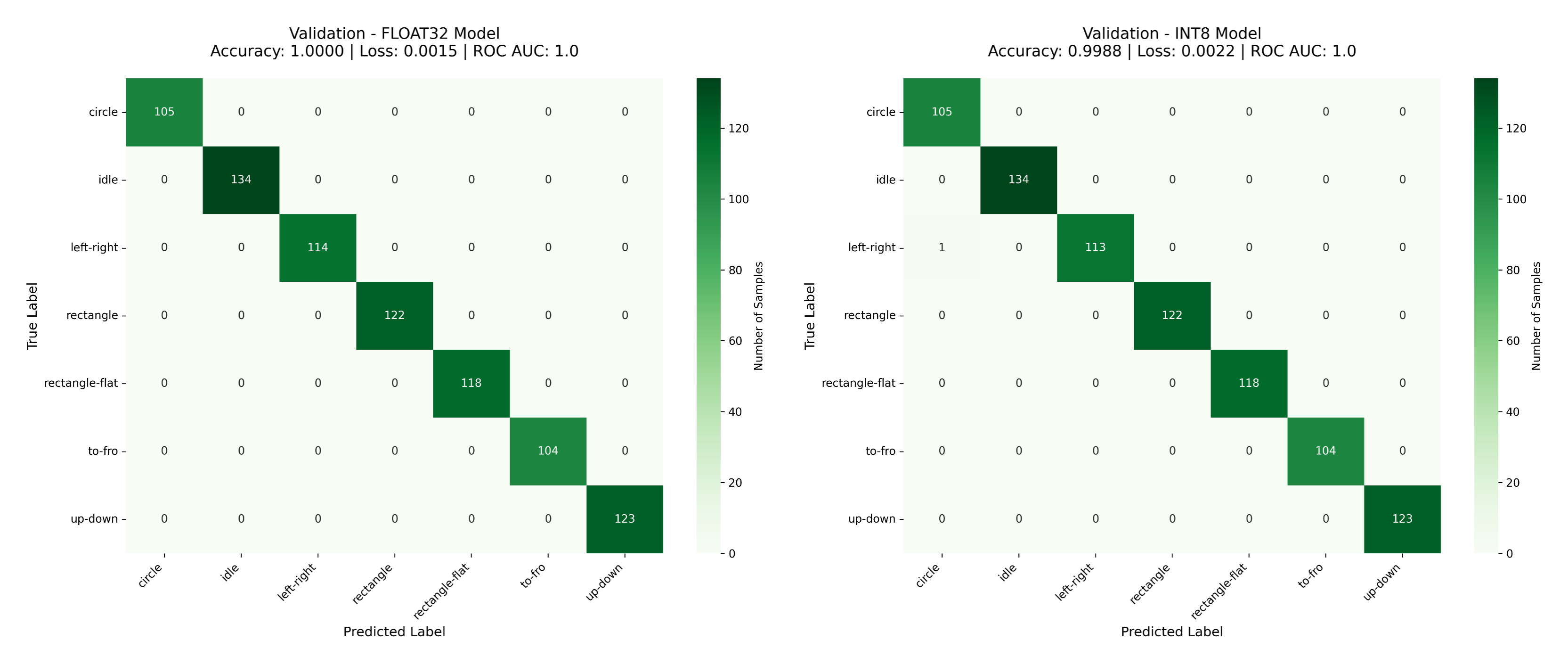}
    \end{adjustbox}
    \caption{Confusion matrix (validation set): float32 vs.\ int8 model.}
    \label{fig:val-con-matrix}
\end{figure*}

\begin{figure}[!t]
    \centering
    \includegraphics[width=0.9\linewidth]{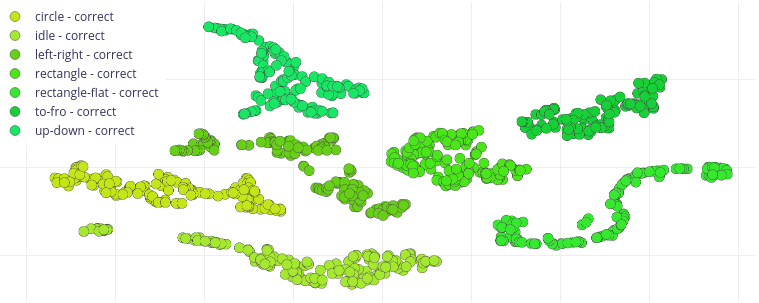}
    \caption{Classified data.}
    \label{fig:dataexpo}
\end{figure}

The classifier is a fully connected network for 117-feature inputs: Dense(20)–Dense(10)–Dense(5) with nonlinear activations, and a 7-unit Softmax output. Trained for 30 epochs using a learned optimizer (base LR \(=5\times10^{-4}\)), batch size 32, and 20\% validation split. The float32 model achieved validation accuracy \(=1.0000\) (loss \(=0.0015\)); the int8-quantized model achieved \(0.9988\) (loss \(=0.0022\)). The quantized model was compiled with the Edge Impulse Neural (EON) compiler and deployed on the Nano 33 BLE Sense. Fig.~\ref{fig:val-con-matrix} shows the validation confusion matrices; Fig.~\ref{fig:dataexpo} shows training-set classifications. On-device metrics are summarized in Table~\ref{tab:performance-table}.

\begin{table}[!t]
    \centering
    \caption{On-Device Performance Metrics of the Compressed NN Model}
    \label{tab:performance-table}
    \begin{tabular}{ll}
        \toprule
        \textbf{Metrics} & \textbf{Result} \\
        \midrule
        Inferencing Time & 1 ms \\
        Peak RAM Usage   & 1.7 KB \\
        Flash Usage      & 17.9 KB \\
        \bottomrule
    \end{tabular}
\end{table}

\vspace{0.3cm}

\subsubsection{Manipulator Visualization}
Gesture-driven motion of the Kinova Gen3 is visualized in RViz within ROS~2 Jazzy. A launch file starts a service node that uses \texttt{MoveGroupInterface} to execute six predefined poses/trajectories, each bound to a gesture. Joint angles are specified in radians (e.g., homing derived from \(130^\circ\times\pi/180\)); velocity and acceleration scaling are 0.1 and 0.05, respectively. The KDL IK solver runs with a \(0.2~\text{s}\) timeout. 

A Python script reads Nano 33 BLE Sense serial data at 115200~baud. The onboard 7-class neural network fuses accelerometer, gyroscope, and magnetometer data; gestures \{idle, up-down, to-fro, left-right, rectangle, rectangle-flat, circle\} are mapped to service calls. The mapping is in Table~\ref{tab:gesture-mapping-manip}. End-to-end latency from gesture to motion starts is \(<100~\text{ms}\), where the TinyML inference \(<50~\text{ms}\) and serial \(<10~\text{ms}\).

\begin{figure*}[!t]
    \centering
    \begin{adjustbox}{max width=\textwidth, center}
        \includegraphics{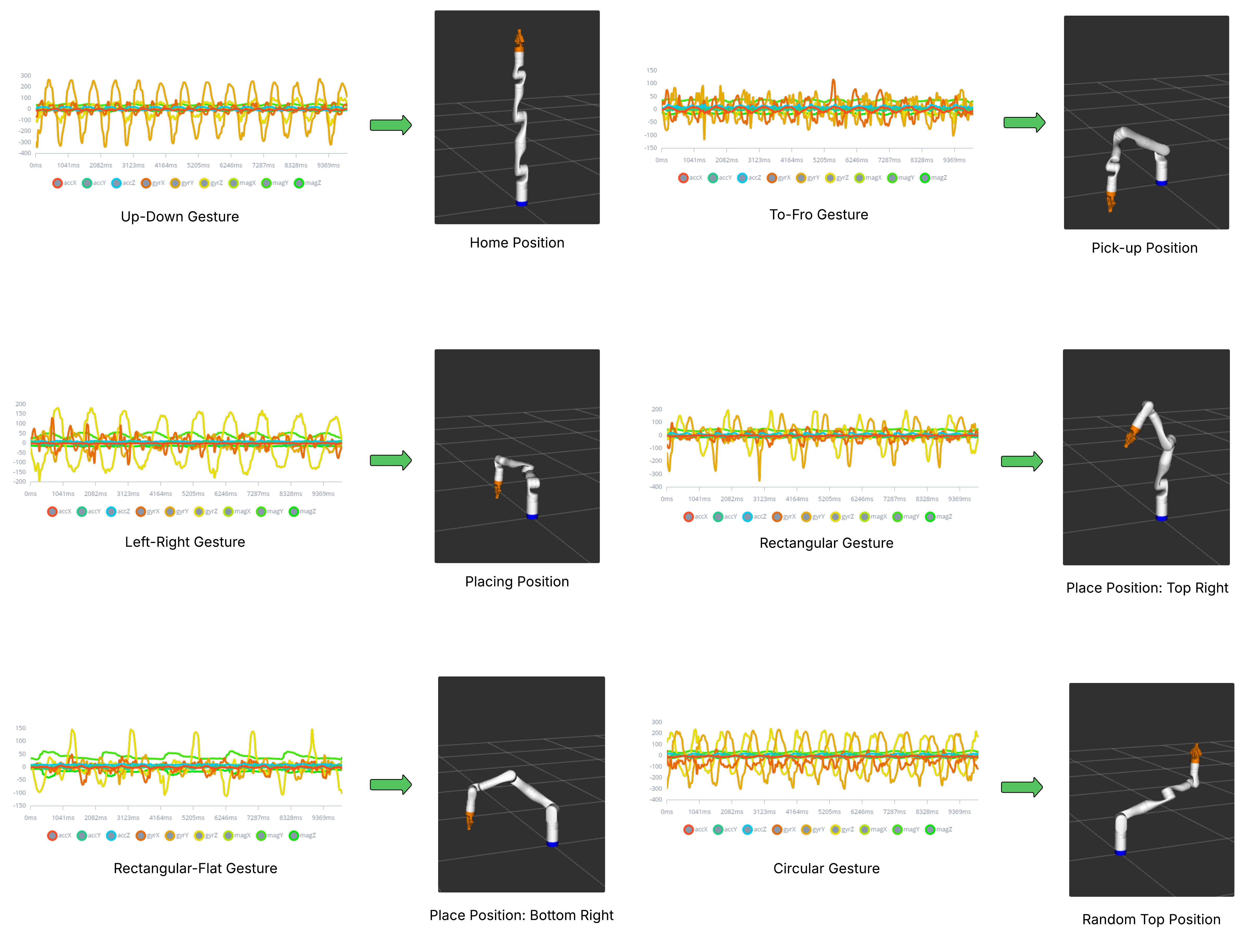}
    \end{adjustbox}
    \caption{Manipulator visualization in RViz executing gestures.}
    \label{fig:manip-vis}
\end{figure*}

RViz shows the arm pose, planned trajectories, and end-effector paths; collision avoidance uses MoveIt~2’s planning scene. Fig.~\ref{fig:manip-vis} illustrates execution.

\begin{table}[!t]
    \centering
    \caption{Gesture-to-Action Mapping for Manipulator Control}
    \label{tab:gesture-mapping-manip}
    \begin{tabular}{ll}
        \toprule
        \textbf{Gesture} & \textbf{Manipulator Action} \\
        \midrule
        up-down        & Homing position \\
        to-fro         & Pick-up trajectory \\
        left-right     & Place trajectory \\
        rectangle      & Place trajectory-2 (Top-right) \\
        rectangle-flat & Place trajectory-3 (Bottom-right) \\
        circle         & Random top trajectory \\
        \bottomrule
    \end{tabular}
\end{table}

\section{Experiments and Results}\label{sec:results}

\subsection{Gesture Recognition Performance}
The gesture recognition system was evaluated on a test set comprising a balanced mixture of pre-recorded data from the original gesture dataset and real-time gestures performed by different users. Each gesture class (\textit{up-down}, \textit{to-fro}, \textit{left-right}, \textit{rectangle}, \textit{rectangle-flat}, \textit{circle}, and \textit{idle}) was tested with variable repetitions per user Testing was conducted in real time to validate robustness under dynamic conditions, including variations in gesture speed and orientation.

\begin{figure*}[!t]
    \centering
    \begin{adjustbox}{max width=\textwidth, center}
        \includegraphics{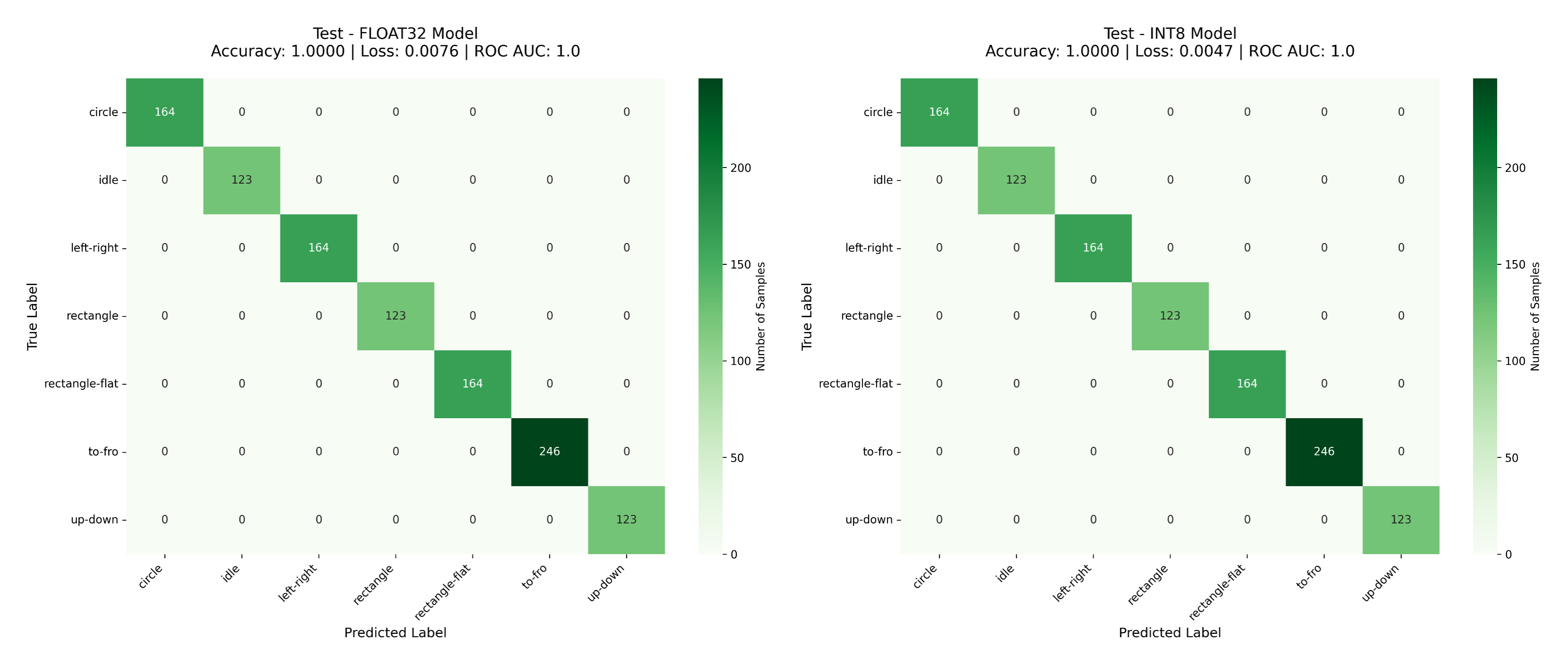}
    \end{adjustbox}
    \caption{Confusion matrix (test set): float32 vs.\ int8 model.}
    \label{fig:confusion-matrices}
\end{figure*}

Both the unquantized float32 and quantized int8 models achieved near-perfect accuracy on the test set, with almost no misclassifications across all gesture classes. Moreover, the absence of false positives in the \textit{idle} class further demonstrates reliability in rejecting unintentional movements. This classification is reflected in the confusion matrices in Fig.~\ref{fig:confusion-matrices}, where almost all predictions align with the ground-truth labels for both models. Notably, int8 quantization preserved classification fidelity while reducing resource demands. The EON compiler produced highly optimized C++ for embedded deployment, reducing RAM usage by 15\% (from 2.0~KB to 1.7~KB) and flash by 19\% (from 22.1~KB to 17.9~KB) compared to the float32 model. Table~\ref{tab:model-performance} summarizes latency, RAM, and flash memory.

The int8 model exhibited a 1~ms inference latency (vs.\ 2~ms for float32), ensuring real-time responsiveness with a 50\% reduction in compute time. Peak RAM decreased from 2.0~KB to 1.7~KB (15\%), and flash from 22.1~KB to 17.9~KB (19\%). The spectral feature extraction pipeline remained a significant contributor to total runtime, but its relative impact was mitigated by the smaller classifier footprint post-quantization. These results highlight TinyML’s ability to balance efficiency and accuracy on resource-constrained devices such as the Arduino Nano 33 BLE Sense.

\begin{table}[!t]
    \centering
    \caption{Performance Metrics of Float32 and Int8 Models}
    \label{tab:model-performance}
    \begin{tabular}{lccc}
        \toprule
        \textbf{Model} & \textbf{Latency (ms)} & \textbf{RAM (KB)} & \textbf{Flash (KB)} \\
        \midrule
        Float32 & 2 & 2.0 & 22.1 \\
        Int8    & 1 & 1.7 & 17.9 \\
        \bottomrule
    \end{tabular}
\end{table}

\subsection{Mobile Base Control Evaluation}\label{subsec:mobile-eval}
The mobile base control was evaluated for directional accuracy and speed modulation using the Arduino-based sensor setup in Sec.~\ref{sec:methodology}, within a ROS2--Gazebo environment. We assessed (i) directional control via the ADXL335 accelerometer and (ii) speed adjustments via flex sensors. For directional control, tilt thresholds of \(\pm 15^\circ\) were set for forward, backward, left, and right commands. Across 400 trials (100 per direction), tilt angles \((\theta\) for X-axis, \(\phi\) for Y-axis\()\) were mapped to \texttt{/cmd\_vel} Twist messages. For speed control, two flex sensors provided \(\pm 0.05~\si{m/s}\) steps over 0 to \(0.5~\si{m/s}\), with 100 trials per sensor for threshold detection plus responsiveness/stability tests.

\begin{figure*}[!t]
    \centering
    \begin{adjustbox}{max width=0.55\textwidth, center}
        \includegraphics{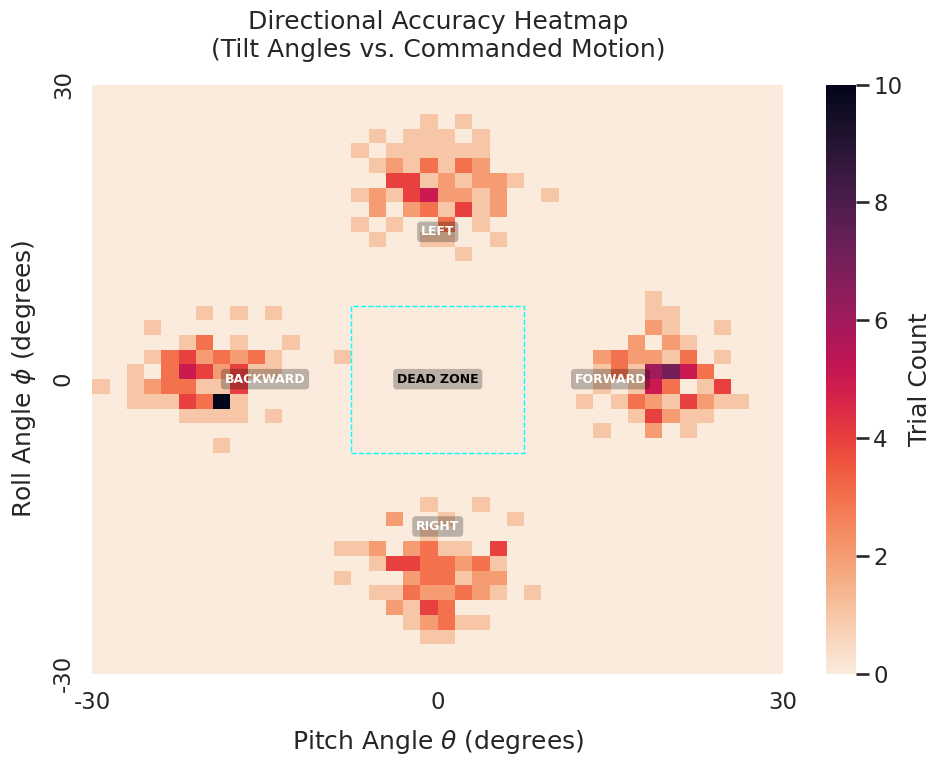}
    \end{adjustbox}
    \caption{Directional accuracy heatmap (tilt angles vs.\ commanded motion).}
    \label{fig:directional-heatmap}
\end{figure*}

Directional accuracy reached 93.5\% across 400 trials, with 6.5\% false triggers attributed to transient noise, mitigated by a neutral dead zone \((|\theta|,|\phi|<5^\circ)\). Forward commands \((\theta>+15^\circ)\) succeeded in 96\% of trials with \(120\pm15~\si{ms}\) response; backward \((\theta<-15^\circ)\) achieved 94\% at \(125\pm20~\si{ms}\); left turns \((\phi>+15^\circ)\) achieved 91\% at \(130\pm25~\si{ms}\); right turns \((\phi<-15^\circ)\) achieved 93\% at \(115\pm10~\si{ms}\) (tilt detection to Gazebo motion). Fig.~\ref{fig:directional-heatmap} visualizes the tilt-to-motion mapping.

Speed modulation was reliable: accelerator (index) detections were 98\% and
decelerator (middle) 95\%, with $<2\%$ false positives. Each bend adjusted the
\emph{maximum} linear and angular speed caps by $\pm 10\%$ (index $+10\%$, middle $-10\%$).
Starting from $v_{\max,0}=0.50~\mathrm{m/s}$ and clipping at $v_{\max}^{\max}=1.00~\mathrm{m/s}$,
reaching the upper cap required $8$ index bends
($0.50\times 1.1^{8}\approx 1.07~\mathrm{m/s}\rightarrow 1.00~\mathrm{m/s}$); reducing from the
upper cap to $\approx 0.48~\mathrm{m/s}$ took $7$ middle bends
($1.0\times 0.9^{7}\approx 0.48~\mathrm{m/s}$).
As summarized in Table~\ref{tab:speed-modulation}, a representative 10~s profile comprised:
1.5~s idle, successive $+10\%$ updates until the cap saturated at $1.00~\mathrm{m/s}$,
3~s steady at the cap, successive $-10\%$ updates, and a final 3~s hold.
The commanded speed tracked the evolving caps with steady-state drift $<0.01~\mathrm{m/s}$.

Overall, the \(\pm 15^\circ\) thresholds and dead zone were validated for robust directional control; flex thresholds yielded 95--98\% detection. Mean command latency for direction and speed remained \(<150~\si{ms}\) (avg.\ \(140\pm30~\si{ms}\)); occasional rapid successive bends caused minor queuing delays.

\begin{table}[!t]
    \centering
    \caption{Speed Modulation Over 10~s with \(\pm10\%\) Cap Adjustments}
    \label{tab:speed-modulation}
    \begin{tabular}{lcc}
        \toprule
        \textbf{Phase} & \textbf{Time (s)} & \textbf{Speed (\si{m/s})} \\
        \midrule
        Initial Hold            & 0.0--1.5  & \(0.00 \pm 0.01\) \\
        Cap Increase (+10\%/bend)   & 1.5--3.0  & \(0.50 \rightarrow 1.00 \pm 0.01\) \\
        Steady at Cap           & 3.0--6.0  & \(1.00 \pm 0.01\) \\
        Cap Decrease (--10\%/bend)  & 6.0--7.5  & \(1.00 \rightarrow 0.50 \pm 0.01\) \\
        Final Hold              & 7.5--10.0 & \(0.00 \pm 0.01\) \\
        \bottomrule
    \end{tabular}

    \vspace{2pt}
    \footnotesize
    Index bends (\texttt{q}) increase, and middle bends (\texttt{z}) decrease the
    maximum speed by \(10\%\) per event. Here \(v_{\max,0}=0.50~\si{m/s}\) and
    \(v_{\max}^{\max}=1.00~\si{m/s}\); the commanded speed follows the evolving cap.
\end{table}

\subsection{Manipulator Trajectory Execution}\label{subsec:manip-eval}
We evaluated gesture-driven trajectories of the Kinova Gen3 to assess real-time performance, precision, and safety within ROS2, using the TinyML recognizer and MoveIt2 integration in Sec.~\ref{sec:methodology}. The Nano 33 BLE Sense model classified six gestures (\textit{up-down}, \textit{to-fro}, \textit{left-right}, \textit{rectangle}, \textit{rectangle-flat}, \textit{circle}), each mapped to a predefined joint-space trajectory executed by MoveIt2. Tests focused on motion control (no end-effector tasks), measuring gesture-to-action latency, joint positioning precision, and collision avoidance in cluttered scenes. A notable constraint emerged: direct planning between arbitrary poses frequently failed unless returning to a home pose first, indicating kinematic/planning limitations.

End-to-end latency from gesture completion to motion start averaged \(110\pm15~\si{ms}\): \(\approx 1~\si{ms}\) TinyML inference and \(90\pm10~\si{ms}\) for service dispatch to MoveIt~2, with the remainder in command handling. Joint positioning showed negligible end-point error across trials. Table~\ref{tab:task-performance} summarizes per-gesture execution times.

The home-pose dependency stems from Inverse Kinematics (IK)/planning challenges in a 7-DOF arm; homing resets to a known collision-free state, simplifying IK and reducing planner failures. With Open Motion Planning Library's (OMPL) default settings, long-range planning suffered from sampling/collision-check costs; segmenting motions via home avoided singularities and respected conservative velocity/acceleration scaling (0.1/0.05). This yields high reliability (100\% collision avoidance in our tests) at the cost of additional homing time. Future improvements could include longer planning times, alternative planners (e.g., RRTConnect, CHOMP), intermediate waypoints, or relaxed jump thresholds.

Overall, sub-second responsiveness and repeatable joint accuracy meet real-time HRI needs. While homing adds overhead, it provided robust execution in our setup.

\begin{table}[!t]
    \centering
    \caption{Gesture-Driven Trajectory Performance}
    \label{tab:task-performance}
    \begin{tabular}{>{\raggedright}p{2cm}p{5cm}p{1.2cm}}
        \toprule
        \textbf{Gesture} & \textbf{Manipulator Action} & \textbf{Avg.\ Execution Time (s)} \\
        \midrule
        Up-Down        & Homing position                     & 3.3 \\
        To-Fro         & Pick-up trajectory                   & 5.7 \\
        Left-Right     & Place trajectory                     & 6.1 \\
        Rectangle      & Place trajectory-2 (top-right)       & 6.4 \\
        Rectangle-Flat & Place trajectory-3 (bottom-right)    & 7.1 \\
        Circle         & Random top trajectory                 & 5.9 \\
        \bottomrule
    \end{tabular}
\end{table}

\section{Discussion}\label{sec:discussion}
The proposed dual-hand gesture control system for ROS-based mobile manipulators advances human-robot interaction (HRI) by integrating TinyML optimization, sensor fusion robustness, and bimanual coordination to tackle challenges in intuitiveness, efficiency, and scalability. This section reflects on experimental outcomes, contrasts them with prior work, and explores the system’s strengths, limitations, and future potential, situating it within the broader landscape of robotic control.

The bimanual architecture leverages natural human motor patterns—left-hand coarse navigation via tilt and flex sensors, right-hand precise manipulator control via TinyML—reducing cognitive load and enabling simultaneous tasks, unlike single-hand systems such as Yu et al. \cite{b1} or Solly et al. \cite{b11}. This contrasts with EMG-driven setups (e.g., Cruz et al. \cite{b3}) hampered by fatigue or vision-based systems (e.g., Altayeb \cite{b2}) limited by camera fields, offering an edge in applications like search-and-rescue or industrial automation where parallel operation is key. TinyML delivered 99.9\% accuracy and 1 ms latency for the neural network, with int8 quantization slashing flash usage by 19\% (22.1 KB to 17.9 KB), though FFT-based features restricted recognition to periodic gestures, sacrificing adaptability for static poses compared to heavier DNNs \cite{b10}. The compact model suits low-cost platforms, but retraining for new gestures remains off-device. ROS2’s modularity enabled seamless integration of mobile base and manipulator nodes, supporting scalability across platforms (e.g., TurtleBot4), yet its 90 ms MoveIt2 latency may bottleneck sub-50 ms applications. This blend of bimanual intuition, edge-AI efficiency, and ROS flexibility outperforms prior art in coordination and resource use, though it balances adaptability against simplicity.

The system’s reliance on predefined manipulator trajectories restricts adaptability to dynamic scenes, as the "pickup" gesture targets fixed poses regardless of object location, while the Nano’s 10 m Bluetooth range limits operation in expansive settings like warehouses, and MoveIt2’s home pose dependency adds a 3.3 s overhead per cycle due to IK singularities. Gesture consistency assumes uniform user input, yet informal multi-user tests can drop accuracy due to hand size or speed variations, and while sensor fusion mitigates noise, extreme vibrations or magnetic interference could disrupt IMU data in harsh environments. The Arduino Nano’s 32 KB SRAM caps model complexity at seven gesture classes, constraining vocabulary expansion without hardware upgrades. These factors highlight a design favoring efficiency and low cost over full autonomy or robustness in unconstrained scenarios, suggesting areas for refinement.

Future enhancements can build on the system’s foundation to broaden its scope and utility:
\begin{itemize}
    \item \textbf{Real-World Deployment}: Port to a physical Kinova Gen3 and Husky rover to test robustness under real noise, terrain, and payload dynamics.
    \item \textbf{Haptic Feedback}: Integrate vibrotactile gloves (e.g., 5 Hz for success, 20 Hz for collisions) to enhance situational awareness beyond RViz visuals.
    \item \textbf{Autonomous Augmentation with SLAM}: Add SLAM (e.g., RTAB-Map) for hybrid control, where gestures trigger autonomous waypoint navigation.
    \item \textbf{Gesture Personalization}: Use federated learning to retrain the TinyML model on-device, adapting to user-specific styles while preserving privacy.
    \item \textbf{Dynamic Motion Planning}: Replace fixed trajectories with real-time planners (e.g., RRTConnect, CHOMP) to eliminate homing, balancing latency and safety.
    \item \textbf{Multi-Modal Inputs}: Pair gestures with voice or eye-tracking (e.g., "grasp" via voice) for finer control granularity.
    \item \textbf{Energy Optimization}: Employ solar-powered wearables or piezoelectric flex sensors to extend uptime in remote settings.
    \item \textbf{Robustness Enhancements}: Fuse IMU with barometric or ultrasonic sensors to counter environmental interference in harsh conditions.
    \item \textbf{Real-Time Gesture-Based Object Manipulation in VR/AR}: Integrate with VR/AR (e.g., Unity + ROS) for 3D virtual object control, aiding training or teleoperation.
    \item \textbf{Wearable Exoskeleton Glove for Haptic Feedback and Force Sensing}: Upgrade to an exoskeleton with actuators and force sensors for bidirectional feedback and grip precision.
    \item \textbf{Custom PCB for Compact Professional Design}: Develop a custom PCB to shrink the wearable footprint, boost durability, and add sensors (e.g., pressure).
    \item \textbf{Hybrid Controller with Voice Commands}: Combine gestures with voice (e.g., "stop" via microphone + ROS) for inclusive, hands-free control.
    \item \textbf{Gesture Control with Biofeedback}: Incorporate biosensors (e.g., heart rate via PPG, skin temperature) to adapt sensitivity to user stress or fatigue.
\end{itemize}

The system’s low-cost, open-source nature fosters accessibility, potentially empowering small enterprises, developing regions, or DIY robotics communities. Its high accuracy and low latency rival commercial systems like Leap Motion, yet at reduced cost and power, aligning with sustainable robotics trends. Ethically, it supports inclusive HRI—e.g., aiding stroke survivors, but raises concerns about over-reliance on automation in critical tasks, necessitating human-in-the-loop safeguards. By integrating bimanual intuition with edge-AI efficiency, this framework sets a foundation for next-generation mobile manipulation, with future enhancements poised to amplify its impact across diverse applications.

\section{Conclusion}\label{sec:conclusion}
This study presented a dual-hand gesture control system for ROS-based mobile manipulators, integrating TinyML, sensor fusion, and a bimanual architecture to advance human-robot interaction (HRI). By employing left-hand tilt and flex sensors for mobile base navigation and right-hand IMU-driven TinyML classification for manipulator control, the system achieved a 99.9\% gesture recognition accuracy with a 1 ms inference latency for the neural network, alongside 93.5\% directional accuracy and precise trajectory execution in a ROS2-Gazebo simulation environment. These results mark a significant improvement over prior single-hand, vision-based, or EMG-driven approaches, delivering an intuitive, low-power interface that excels in efficiency and coordination, as validated against related work in Section~\ref{sec:related_work}. Our core contributions include: (1) a dual-hand control architecture enabling simultaneous navigation and manipulation, (2) a TinyML-based system for real-time, low-power gesture recognition, (3) a sensor fusion approach ensuring robust mobile base control, and (4) a comprehensive ROS2 simulation framework unifying hardware and software layers. This framework’s open-source, cost-effective design extends its potential to industrial automation, assistive robotics, and hazardous operations, offering a scalable solution for diverse applications. While limitations such as predefined trajectories and Bluetooth range persist, the system lays a strong foundation for next-generation HRI, with future enhancements poised to further its impact in dynamic, real-world settings.

\bibliographystyle{IEEEtran}
\bibliography{references}

\begin{thebibliography}{10}
\providecommand{\url}[1]{#1}
\csname url@samestyle\endcsname
\providecommand{\newblock}{\relax}
\providecommand{\bibinfo}[2]{#2}
\providecommand{\BIBentrySTDinterwordspacing}{\spaceskip=0pt\relax}
\providecommand{\BIBentryALTinterwordstretchfactor}{4}
\providecommand{\BIBentryALTinterwordspacing}{\spaceskip=\fontdimen2\font plus
\BIBentryALTinterwordstretchfactor\fontdimen3\font minus \fontdimen4\font\relax}
\providecommand{\BIBforeignlanguage}[2]{{%
\expandafter\ifx\csname l@#1\endcsname\relax
\typeout{** WARNING: IEEEtran.bst: No hyphenation pattern has been}%
\typeout{** loaded for the language `#1'. Using the pattern for}%
\typeout{** the default language instead.}%
\else
\language=\csname l@#1\endcsname
\fi
#2}}
\providecommand{\BIBdecl}{\relax}
\BIBdecl

\bibitem{spacerobo01}
Y.~Gao and S.~Chien, ``Review on space robotics: Toward top-level science through space exploration,'' \emph{Science Robotics}, vol.~2, no.~7, p. eaan5074, 2017.

\bibitem{spacerobo02}
Y.~Xu and T.~Kanade, \emph{Space robotics: dynamics and control}.\hskip 1em plus 0.5em minus 0.4em\relax Springer Science \& Business Media, 1992, vol. 188.

\bibitem{spacerobo03}
D.~S. Katz and R.~R. Some, ``Nasa advances robotic space exploration,'' \emph{Computer}, vol.~36, no.~1, pp. 52--61, 2003.

\bibitem{disasterrobo01}
S.~Park, Y.~Oh, and D.~Hong, ``Disaster response and recovery from the perspective of robotics,'' \emph{International Journal of Precision Engineering and Manufacturing}, vol.~18, pp. 1475--1482, 2017.

\bibitem{disasterrobo02}
G.~Wilk-Jakubowski, R.~Harabin, and S.~Ivanov, ``Robotics in crisis management: A review,'' \emph{Technology in Society}, vol.~68, p. 101935, 2022.

\bibitem{militaryrobo01}
M.~Sangeetha and K.~Srinivasan, ``Swarm robotics: a new framework of military robots,'' in \emph{Journal of Physics: Conference Series}, vol. 1717, no.~1.\hskip 1em plus 0.5em minus 0.4em\relax IOP Publishing, 2021, p. 012017.

\bibitem{militaryrobo02}
B.~Choi, W.~Lee, G.~Park, Y.~Lee, J.~Min, and S.~Hong, ``Development and control of a military rescue robot for casualty extraction task,'' \emph{Journal of Field Robotics}, vol.~36, no.~4, pp. 656--676, 2019.

\bibitem{assrobo01}
M.~J. Matari{\'c} and B.~Scassellati, ``Socially assistive robotics,'' \emph{Springer handbook of robotics}, pp. 1973--1994, 2016.

\bibitem{assrobo02}
S.~H. Bengtson, T.~Bak, L.~N. Andreasen~Struijk, and T.~B. Moeslund, ``A review of computer vision for semi-autonomous control of assistive robotic manipulators (arms),'' \emph{Disability and Rehabilitation: Assistive Technology}, vol.~15, no.~7, pp. 731--745, 2020.

\bibitem{neuro_arm}
G.~R. Sutherland, S.~Wolfsberger, S.~Lama, and K.~Zarei-nia, ``The evolution of neuroarm,'' \emph{Neurosurgery}, vol.~72, pp. A27--A32, 2013.

\bibitem{momaindustry01}
N.~Ghodsian, K.~Benfriha, A.~Olabi, V.~Gopinath, and A.~Arnou, ``Mobile manipulators in industry 4.0: A review of developments for industrial applications,'' \emph{Sensors}, vol.~23, no.~19, p. 8026, 2023.

\bibitem{momaindustry02}
F.~Chen, B.~Gao, M.~Selvaggio, Z.~Li, D.~Caldwell, K.~Kershaw, A.~Masi, M.~Di~Castro, and R.~Losito, ``A framework of teleoperated and stereo vision guided mobile manipulation for industrial automation,'' in \emph{2016 IEEE International Conference on Mechatronics and Automation}.\hskip 1em plus 0.5em minus 0.4em\relax IEEE, 2016, pp. 1641--1648.

\bibitem{momahealth01}
Z.~Li, P.~Moran, Q.~Dong, R.~J. Shaw, and K.~Hauser, ``Development of a tele-nursing mobile manipulator for remote care-giving in quarantine areas,'' in \emph{2017 IEEE International Conference on Robotics and Automation (ICRA)}.\hskip 1em plus 0.5em minus 0.4em\relax IEEE, 2017, pp. 3581--3586.

\bibitem{momahealth02}
J.~Varela-Ald{\'a}s, J.~Buele, S.~Guerrero-N{\'u}{\~n}ez, and V.~H. Andaluz, ``Mobile manipulator for hospital care using firebase,'' in \emph{International Conference on Human-Computer Interaction}.\hskip 1em plus 0.5em minus 0.4em\relax Springer, 2022, pp. 328--341.

\bibitem{momasearch01}
F.~Pastor, F.~J. Ruiz-Ruiz, J.~M. G{\'o}mez-de Gabriel, and A.~J. Garc{\'\i}a-Cerezo, ``Autonomous wristband placement in a moving hand for victims in search and rescue scenarios with a mobile manipulator,'' \emph{IEEE Robotics and Automation Letters}, vol.~7, no.~4, pp. 11\,871--11\,878, 2022.

\bibitem{coal}
L.~Tong, M.~Zhang, H.~Ma, C.~Wang, and L.~Peng, ``Semg-based gesture recognition method for coal mine inspection manipulator using multistream cnn,'' \emph{IEEE Sensors Journal}, vol.~23, no.~10, pp. 11\,082--11\,090, 2023.

\bibitem{bomb}
M.~Shyam, M.~Amalasweena, K.~Balasaranya, R.~Renugadevi, K.~P. Chandran \emph{et~al.}, ``Intellectual design of bomb identification and defusing robot based on logical gesturing mechanism,'' in \emph{2023 International Conference on Advances in Computing, Communication and Applied Informatics (ACCAI)}.\hskip 1em plus 0.5em minus 0.4em\relax IEEE, 2023, pp. 1--8.

\bibitem{nuclear}
A.~E. Salman and M.~R. Roman, ``Augmented reality-assisted gesture-based teleoperated system for robot motion planning,'' \emph{Industrial Robot: the international journal of robotics research and application}, vol.~50, no.~5, pp. 765--780, 2023.

\bibitem{visiongesture01}
J.~Qi, L.~Ma, Z.~Cui, and Y.~Yu, ``Computer vision-based hand gesture recognition for human-robot interaction: a review,'' \emph{Complex \& Intelligent Systems}, vol.~10, no.~1, pp. 1581--1606, 2024.

\bibitem{visiongesture02}
N.~Mohamed, M.~B. Mustafa, and N.~Jomhari, ``A review of the hand gesture recognition system: Current progress and future directions,'' \emph{IEEE access}, vol.~9, pp. 157\,422--157\,436, 2021.

\bibitem{gestureDNN}
P.~J. Cruz, J.~P. V{\'a}sconez, R.~Romero, A.~Chico, M.~E. Benalc{\'a}zar, R.~{\'A}lvarez, L.~I. Barona~L{\'o}pez, and {\'A}.~L. Valdivieso~Caraguay, ``A deep q-network based hand gesture recognition system for control of robotic platforms,'' \emph{Scientific Reports}, vol.~13, no.~1, p. 7956, 2023.

\bibitem{gestureCNN}
Z.~Yu, C.~Lu, Y.~Zhang, and L.~Jing, ``Gesture-controlled robotic arm for agricultural harvesting using a data glove with bending sensor and optitrack systems,'' \emph{Micromachines}, vol.~15, no.~7, p. 918, 2024.

\bibitem{stroke_rehab}
J.~G. Colli-Alfaro, A.~Ibrahim, and A.~L. Trejos, ``Design of user-independent hand gesture recognition using multilayer perceptron networks and sensor fusion techniques,'' in \emph{2019 IEEE 16th International Conference on Rehabilitation Robotics (ICORR)}.\hskip 1em plus 0.5em minus 0.4em\relax IEEE, 2019, pp. 1103--1108.

\bibitem{emg}
E.~Kim, J.~Shin, Y.~Kwon, and B.~Park, ``Emg-based dynamic hand gesture recognition using edge ai for human--robot interaction,'' \emph{Electronics}, vol.~12, no.~7, p. 1541, 2023.

\bibitem{wso_stroke}
V.~L. Feigin, M.~Brainin, B.~Norrving, S.~Martins, R.~L. Sacco, W.~Hacke, M.~Fisher, J.~Pandian, and P.~Lindsay, ``World stroke organization (wso): global stroke fact sheet 2022,'' \emph{International Journal of Stroke}, vol.~17, no.~1, pp. 18--29, 2022.

\bibitem{rw_control}
B.~Sabuj, M.~J. Islam, and M.~A. Rahaman, ``Human robot interaction using sensor based hand gestures for assisting disable people,'' in \emph{2019 International Conference on Sustainable Technologies for Industry 4.0 (STI)}.\hskip 1em plus 0.5em minus 0.4em\relax IEEE, 2019, pp. 1--5.

\bibitem{tinymlintro01}
P.~P. Ray, ``A review on tinyml: State-of-the-art and prospects,'' \emph{Journal of King Saud University-Computer and Information Sciences}, vol.~34, no.~4, pp. 1595--1623, 2022.

\bibitem{tinymlintro02}
Y.~Abadade, A.~Temouden, H.~Bamoumen, N.~Benamar, Y.~Chtouki, and A.~S. Hafid, ``A comprehensive survey on tinyml,'' \emph{IEEE Access}, vol.~11, pp. 96\,892--96\,922, 2023.

\bibitem{tinymlintro03}
N.~Schizas, A.~Karras, C.~Karras, and S.~Sioutas, ``Tinyml for ultra-low power ai and large scale iot deployments: A systematic review,'' \emph{Future Internet}, vol.~14, no.~12, p. 363, 2022.

\bibitem{sensefus01}
S.~Majumder and N.~Kehtarnavaz, ``Vision and inertial sensing fusion for human action recognition: A review,'' \emph{IEEE Sensors Journal}, vol.~21, no.~3, pp. 2454--2467, 2020.

\bibitem{sensefus02}
R.~Tchantchane, H.~Zhou, S.~Zhang, and G.~Alici, ``A review of hand gesture recognition systems based on noninvasive wearable sensors,'' \emph{Advanced intelligent systems}, vol.~5, no.~10, p. 2300207, 2023.

\bibitem{b1}
Z.~Yu, C.~Lu, Y.~Zhang, and L.~Jing, ``Gesture-controlled robotic arm for agricultural harvesting using a data glove with bending sensor and optitrack systems,'' \emph{Micromachines}, vol.~15, no.~7, p. 918, 2024.

\bibitem{b6}
M.~M.~U. Faiz and M.~F. Alshammari, ``Implementation of a wireless human hand gesture controlled robotic arm,'' in \emph{2024 IEEE International Conference on Artificial Intelligence and Mechatronics Systems (AIMS)}.\hskip 1em plus 0.5em minus 0.4em\relax IEEE, 2024, pp. 1--6.

\bibitem{b7}
M.~M.~U. Saleheen, M.~R. Fahad, and R.~Khan, ``Gesture-controlled robotic arm,'' in \emph{2023 International Conference on Computer Science, Information Technology and Engineering (ICCoSITE)}.\hskip 1em plus 0.5em minus 0.4em\relax IEEE, 2023, pp. 495--499.

\bibitem{b11}
E.~Solly and A.~Aldabbagh, ``Gesture controlled mobile robot,'' in \emph{2023 5th International Congress on Human-Computer Interaction, Optimization and Robotic Applications (HORA)}, 2023, pp. 1--6.

\bibitem{b2}
M.~Altayeb, ``Hand gestures replicating robot arm based on mediapipe,'' \emph{Indonesian Journal of Electrical Engineering and Informatics (IJEEI)}, vol.~11, no.~3, pp. 727--737, 2023.

\bibitem{b9}
W.~N. Wan~Azlan, W.~N. Wan~Zakaria, N.~Othman, and M.~N.~H. Mohd, ``Development of hand gesture controlled robotic arm for hemiplegia patients,'' in \emph{AIP Conference Proceedings}, vol. 2564, no.~1.\hskip 1em plus 0.5em minus 0.4em\relax AIP Publishing, 2023.

\bibitem{b10}
R.~Mariappan, P.~Gayathri, P.~Pushpalatha, V.~S.~S. Rishik, and T.~Satish, ``Real time robotic arm using leap motion controller,'' in \emph{Journal of Physics: Conference Series}, vol. 2466, no.~1.\hskip 1em plus 0.5em minus 0.4em\relax IOP Publishing, 2023, p. 012023.

\bibitem{b3}
P.~J. Cruz, J.~P. V{\'a}sconez, R.~Romero, A.~Chico, M.~E. Benalc{\'a}zar, R.~{\'A}lvarez, L.~I. Barona~L{\'o}pez, and {\'A}.~L. Valdivieso~Caraguay, ``A deep q-network based hand gesture recognition system for control of robotic platforms,'' \emph{Scientific Reports}, vol.~13, no.~1, p. 7956, 2023.

\bibitem{b8}
E.~Kim, J.~Shin, Y.~Kwon, and B.~Park, ``Emg-based dynamic hand gesture recognition using edge ai for human--robot interaction,'' \emph{Electronics}, vol.~12, no.~7, p. 1541, 2023.

\bibitem{b12}
------, ``Emg-based dynamic hand gesture recognition using edge ai for human--robot interaction,'' \emph{Electronics}, vol.~12, no.~7, p. 1541, 2023.

\bibitem{b5}
K.~Nandhini, C.~Kumar, M.~Prathap, A.~J. Rahamath, and K.~Krishnudu, ``Gesture controlled robotic arm for radioactive environment,'' in \emph{AIP Conference Proceedings}, vol. 2492, no.~1.\hskip 1em plus 0.5em minus 0.4em\relax AIP Publishing, 2023.

\bibitem{b4}
M.~Soori, B.~Arezoo, and R.~Dastres, ``Optimization of energy consumption in industrial robots, a review,'' \emph{Cognitive Robotics}, vol.~3, pp. 142--157, 2023.

\bibitem{pedley2013tilt}
M.~Pedley, ``Tilt sensing using a three-axis accelerometer,'' \emph{Freescale semiconductor application note}, vol.~1, pp. 2012--2013, 2013.

\bibitem{saggio2015resistive}
G.~Saggio, F.~Riillo, L.~Sbernini, and L.~R. Quitadamo, ``Resistive flex sensors: a survey,'' \emph{Smart Materials and Structures}, vol.~25, no.~1, p. 013001, 2015.

\end{thebibliography}

\end{document}